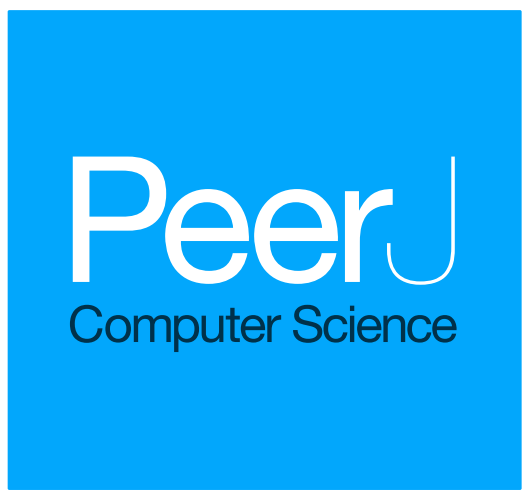

# A syntax-injected approach for faster and more accurate sentiment analysis

Muhammad Imran[1,*], Olga Kellert[1,2,*] and Carlos Gómez-Rodríguez[1]

[1] Universidade da Coruña, CITIC, Departamento de Ciencias de la Computación y Tecnologías de la Información, Campus de Elviña, A Coruña, Spain

[2] School of International Letters and Cultures, The College of Liberal Arts and Sciences, Arizona State University, Tempe, United States

[*] These authors contributed equally to this work.

## ABSTRACT

Sentiment Analysis (SA) is a crucial aspect of Natural Language Processing (NLP), focusing on identifying and interpreting subjective assessments in textual content. Syntactic parsing is useful in SA as it improves accuracy and provides explainability; however, it often becomes a computational bottleneck due to slow parsing algorithms. This article proposes a solution to this bottleneck by using a Sequence Labeling Syntactic Parser (SELSP) to integrate syntactic information into SA *via* a rule-based sentiment analysis pipeline. By reformulating dependency parsing as a sequence labeling task, we significantly improve the efficiency of syntax-based SA. SELSP is trained and evaluated on a ternary polarity classification task, demonstrating greater speed and accuracy compared to conventional parsers like Stanza and heuristic approaches such as Valence Aware Dictionary and sEntiment Reasoner (VADER). The combination of speed and accuracy makes SELSP especially attractive for sentiment analysis applications in both academic and industrial contexts. Moreover, we compare SELSP with Transformer-based models trained on a 5-label classification task. In addition, we evaluate multiple sentiment dictionaries with SELSP to determine which yields the best performance in polarity prediction. The results show that dictionaries accounting for polarity judgment variation outperform those that ignore it. Furthermore, we show that SELSP outperforms Transformer-based models in terms of speed for polarity prediction.

Corresponding authors
Muhammad Imran, m.imran@udc.es
Olga Kellert, olga.kellert@asu.edu

## INTRODUCTION

Sentiment Analysis (SA) holds a pivotal role in Natural Language Processing (NLP), dealing with the identification of subjective assessments, such as opinions on hotels in textual content (subjectivity identification task), and predicting the polarity of these assessments, whether they are positive or negative (Polarity Classification task, PC), as exemplified in (1) (*Cui et al., 2023*).[1]

[1] Portions of this text were previously published as part of a preprint (https://arxiv.org/pdf/2406.15163).

(1) **Anne**: This hotel is awesome. (PC: Positive) **Ben**: This hotel is terrible. (PC: Negative)

Some approaches to SA rely on identifying sentiment-bearing words in a sentence, such as "awesome" and "terrible" in (1), and using them as key indicators of polarity.

These are known as sentiment words, and are listed together with their associated polarities in sentiment dictionaries like SO-CAL (*Brooke, Tofiloski & Taboada, 2009*). Lexicon-based methods like this offer an intuitive, transparent, and easily explainable analysis, as the sentiment polarity associated with a sentence can be traced back to the specific words that triggered it. This stands in contrast to supervised machine learning approaches, which rely on black-box models that infer predictions from training data but do not offer interpretability or word-level traceability (*Cui et al., 2023*).

However, the polarity of a sentiment word can be shifted by various linguistic elements, notably including negation, as in "This hotel is not good." As a consequence, if high accuracy is desired, dictionary-based approaches cannot simply rely on sentiment dictionaries like SO-CAL to perform SA tasks. These approaches should also somehow model the relation between sentiment words and polarity shifting elements like negation. Since the scope of relevant phenomena such as negation can be determined from sentence syntax, syntactic parsing is a natural way to deal with the syntactic relations between sentiment words and polarity-shifting constructs. This is achieved by applying rules that predict sentence polarity based on the structure of a parse tree. These rules offer an explainable and transparent approach and tend to be more accurate than simple local heuristics, such as assigning negation scope to a fixed-length window of tokens (*Thelwall et al., 2010*). The most widely used formalism for syntactic parsing is Universal Dependencies (UD) (*Zeman et al., 2023*). UD is a dependency-based framework that represents the syntactic structure of a sentence as a parse tree composed of directed binary relations (called dependencies) between words. These dependencies specify the syntactic roles that words play in the sentence, such as subject, direct object, or adverbial modifier.

Stanza is an example of a popular NLP tool that provides syntactic dependency parsing in the UD formalism in several languages, including Spanish and English, our focus languages in this article (*Qi et al., 2020*). It has been shown that approaches that use syntactic parsing for an explainable and transparent SA based on the UD formalism or predecessor versions (*Vilares, Gómez-Rodríguez & Alonso, 2017*; *Kellert et al., 2023*) show better performance in polarity prediction tasks than heuristic approaches that do not rely on syntactic parsing at all such as VADER (*Hutto & Gilbert, 2014*), as demonstrated by *Gómez-Rodríguez, Alonso-Alonso & Vilares (2019)*.

Despite the clear advantages of using syntactic parsing for SA, conventional syntactic parsers like Stanza (*Qi et al., 2020*) are often slow and computationally costly (*Gómez-Rodríguez & Vilares, 2018*; *Roca, Vilares & Gómez-Rodríguez, 2023*; *Strzyz, Vilares & Gómez-Rodríguez, 2019b*). Several solutions have been suggested in recent years to address this problem by developing faster systems by the transformation of dependency parsing into a SEquence Labeling (SEL) task (*Gómez-Rodríguez & Vilares, 2018*; *Roca, Vilares & Gómez-Rodríguez, 2023*; *Strzyz, Vilares & Gómez-Rodríguez, 2019b*). However, so far, SEL parsers have not been applied to SA. This article fills this gap. Thus, we adopt the SEL dependency parsing approach for sentiment analysis, to speed up the application of syntactic rules for polarity prediction and evaluate its practicality. In addition, we test our approach with several dictionaries to see which one yields the highest accuracy when used with the SEL-based system.

Our contributions are: (1) We present, to the best of our knowledge, the first approach that performs fast, syntax-based polarity classification using sequence labeling parsing. For this purpose, we first train an SEL system to perform syntactic dependency parsing (SELSP) in English and Spanish. After measuring its parsing accuracy on test sets of these languages, we then apply our SESLP in the domain of SA for the polarity prediction task through a sentiment analysis engine which relies on syntax-based rules to calculate polarity in English and Spanish. (2) We compare the performance of our approach in terms of runtime and accuracy in both languages using the same syntactic rules on top of the well-known dependency parser from Stanza, as well as to VADER—a heuristic approach to SA that does not use syntax trees and is widely used in industry—and to a state-of-the-art Transformer-based (RoBERTa) model. (3) We study the influence of sentiment dictionaries and their interaction with syntax-based systems by using several sentiment dictionaries, and combinations thereof, on top of the same set of rules and the two mentioned parsers.

## RELATED WORK

The most widely used approaches to perform SA include Machine Learning/Deep Learning (ML/DL) methods, lexicon-based methods, and hybrid methods.

### ML/DL methods and hybrid methods

Supervised approaches using machine learning, including deep learning architectures like Long Short-Term Memory (LSTM) or Convolutional Neural Network (CNN) and pre-trained models like BERT and RoBERTa, are a common practice in SA (see *Cui et al., 2023*). These approaches use the standard supervised learning methodology where a training dataset is used during the training process to learn correlations between the specific input text and the sentiment polarity, and the resulting trained system is then evaluated on a separate test dataset and/or used in production. The performance of ML/DL-based approaches is known to be relatively high if the learning and prediction tasks are performed on a similar dataset or *corpus*, as they adapt especially well to the particularities of the training dataset. This means that they often dominate the rankings in shared tasks, such as that of Rest-Mex 2023 (*Álvarez-Carmona et al., 2023*), where training and test datasets are obtained from the same source and exhibit similar characteristics in terms of topic, length, style, *etc*. However, their performance can degrade on texts that differ from the training dataset, even within the same domain (see *e.g*., Table 5 of *Vilares, Gómez-Rodríguez & Alonso (2017)*).

Thus, a downside of these approaches is that they require a large amount of data for learning and prediction, ideally representative of the sentences that the system will see in production. This makes them suboptimal for real-world applications, where such data may be unavailable or costly to acquire, or users may lack the necessary expertise in training and using these models. In addition, the learned associations between input and output are often obscure to be easily understood or explained.

Some approaches, like *Yang et al. (2020)*, use sentiment dictionaries to provide or modify the input information fed to supervised machine learning approaches. Thus, they are often considered hybrid approaches (*Cui et al., 2023*) between ML/DL methods and the lexicon-based methods described in the next section. While exploiting sentiment dictionaries is useful to improve accuracy, their machine-learning-based core means that they still share the drawbacks described above.

## Lexicon-based methods

The lexicon-based methodology makes use of a sentiment dictionary that contains sentiment words with corresponding polarity scores to find out the overall opinion or polarity of a sentence or text (*Taboada et al., 2011*; *Hutto & Gilbert, 2014*; *Vilares, Gómez-Rodríguez & Alonso, 2017*; *Kellert et al., 2023*). These approaches fall into various branches depending on how they deal with syntactic rules that can shift or intensify the polarity. Non-compositional and non-recursive rule-based approaches such as SO-CAL (*Taboada et al., 2011*), SentiStrength (*Thelwall, Buckley & Paltoglou, 2012*) or VADER (*Hutto & Gilbert, 2014*) use heuristic rules to account for polarity changes due to negation, modification, or other syntactic processes. Other unsupervised approaches use a more linguistically-informed architecture to account for SA by exploiting general syntactic rules that influence sentiment polarity, requiring a parse tree (*Tan et al., 2012*; *Vilares, Gómez-Rodríguez & Alonso, 2017*; *Kellert et al., 2023*).

Lexicon-based methods are considered to be simple to use and cost-effective as there is no need to train on huge amounts of data, and they work well across domains. For these reasons, in spite of less attention in the NLP literature in recent years due to the general trend towards machine learning in the field, they still tend to be the most commonly used in industry and production domains such as biomedicine (*e.g*., see Table 5 in *Denecke & Reichenpfader (2023)*), fintech (*Bredice et al., 2025*) or economics (*Algaba et al., 2020*; *Barbaglia et al., 2025*).

To maximize accuracy, lexicon-based methods must reliably determine the scope of relevant syntactic phenomena such as negation or intensification. As mentioned above, the most basic form of scope detection is to use simple local heuristics such as a fixed-length window, but this has limitations in terms of accuracy. For example, systems like SentiStrength or VADER can detect that "good" is negated in "not good", but not in "not in any respect good", due to the long-distance dependency. Instead, one can use machine learning classifiers to determine scope (see *Cruz, Taboada & Mitkov, 2016*; *Jiménez-Zafra et al., 2021* and references therein), but this entails the disadvantages previously discussed for ML approaches, including the need for specific data and lack of explainability. Thus, the alternative on which we focus our proposal is to use syntactic rules that operate on a parse tree (*Jia, Yu & Meng, 2009*; *Vilares, Alonso & Gómez-Rodríguez, 2015*; *Vilares, Gómez-Rodríguez & Alonso, 2017*; *Jimenez Zafra et al., 2019*; *Kellert et al., 2023*). The next sections address methods for syntactic parsing and how they can be used for polarity classification.

## Syntactic parsing for sentiment analysis

Syntactic parsing is the task of determining the hierarchical structure of a sentence in natural language. For this purpose, two main kinds of representations are used in NLP: constituent parsing, where the sentence is recursively divided into smaller units, and dependency parsing, which focuses on representing relations between words. While constituent parsing has seen some use in sentiment analysis (*Socher et al., 2013*), dependency parsing is the dominant alternative for the task, as it straightforwardly represents what words are modified by other words. In particular, it is the basis of all the methods based on syntactic rules cited in the previous section.

In the last few years, Universal Dependencies (UD) (*Zeman et al., 2023*) has become the predominant framework for dependency parsing, especially in multilingual setups, as it provides harmonized annotation guidelines and freely-available corpora for over 100 diverse languages.

Syntactic parsing within UD involves the analysis of the grammatical structure of sentences to determine the roles of each of their parts and the relationships between them. It includes text tokenization that breaks down the input sentence into individual words or tokens, Part-of-Speech (POS) tagging where each token is assigned a tag indicating its grammatical category (*e.g.*, noun, verb, adjective), and finally dependency parsing itself, where the syntactic relationships between words in a sentence are determined. Like all dependency grammar frameworks, UD represents syntactic relationships as directed links between words, where one word is the head and the other is the dependent. The dependency relation typically captures grammatical relations such as subject, object, modifier, *etc*. In UD, the syntactic relations are expressed in a standardized framework for annotating dependency trees for different languages. The output of the syntactic parsing in UD is a dependency tree, which is a hierarchical representation of the syntactic structure of the sentence. Each node in the tree represents a word, and the edges capture the syntactic relationships between words.

*Kellert et al. (2023)* make use of UD-based syntactic parsing and exploit dependency relations expressed in UD between sentiment words and other words that can shift, strengthen, or weaken the polarity of sentiment words (see also *Vilares, Gómez-Rodríguez & Alonso (2017)* for application of syntactic parsing for SA using a predecessor framework to UD). The first stage of the approach in *Kellert et al. (2023)* is to search for sentiment words taken from a sentiment dictionary SO-CAL (*Brooke, Tofiloski & Taboada, 2009*) in

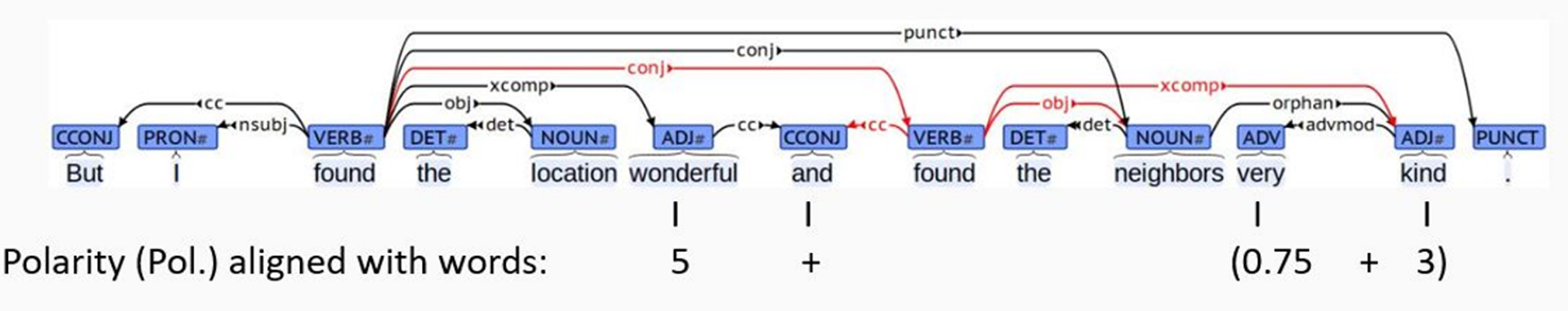

Mean Pol.: (5 + 3.75): 2 = 4.375 ≙ Positive Polarity in a system from 1 to 5, where 5 is positive

```
# sent_id = reviews-044427-0003
# text = But I found the location wonderful and the neighbors very kind.
1    But        but        CCONJ   CC   _                                          3   cc       _           _
2    I          I          PRON    PRP  Case=Nom|Number=Sing|Person=1|PronType=Prs  3   nsubj    _           _
3    found      find       VERB    VBD  Mood=Ind|Tense=Past|VerbForm=Fin            0   root     _           _
4    the        the        DET     DT   Definite=Def|PronType=Art                   5   det      _           _
5    location   location   NOUN    NN   Number=Sing                                 3   obj      _           _
6    wonderful  wonderful  ADJ     JJ   Degree=Pos                                  3   xcomp    _           _
7    and        and        CCONJ   CC   _                                           6   cc       _           _
7.1  found      find       VERB    VBD  Mood=Ind|Tense=Past|VerbForm=Fin            _   _        3:conj      _
8    the        the        DET     DT   Definite=Def|PronType=Art                   9   det      _           _
9    neighbors  neighbor   NOUN    NNS  Number=Plur                                 3   conj     7.1:obj     _
10   very       very       ADV     RB   _                                           11  advmod   _           _
11   kind       kind       ADJ     JJ   Degree=Pos                                  9   orphan   7.1:xcomp   SpaceAfter=No
12   .          .          PUNCT   .    _                                           3   punct    _           _
```

**Figure 1 Polarity prediction based on a UD-based dependency tree from *Kellert et al. (2023)*.** Full-size DOI: 10.7717/peerj-cs.3519/fig-1

the input text, such as the sentiment word *wonderful* (in Fig. 1). Then, the system traverses the dependency tree and applies a set of syntax-based rules to check whether negation words or other modifiers and relevant syntactic phenomena change the polarity of sentiment words (see the dependency tree in Fig. 1). This syntactic approach applies recursive and compositional rules to negation and intensification, to identify the right scope of polarity shifting elements and to properly compute the sentence polarity (*Vilares, Gómez-Rodríguez & Alonso, 2017*; *Kellert et al., 2023*). This means that the computing of the sentence polarity is done bottom-up: first, intensifiers and polarity-shifting syntactic phenomena in the structurally lowest levels of the tree are used to modify the polarity of their modified sentiment-bearing words and expressions. Then, structurally higher polarity shifting elements are computed, and polarities from converging branches of the dependency tree are aggregated, until the root of the sentence structure is reached and the final combined polarity for the whole sentence is obtained (*Vilares, Gómez-Rodríguez & Alonso, 2017*; *Kellert et al., 2023*). For instance, in the sentence in Fig. 1, first, the modification of the adjective "kind" by the adverb "very" is computed, and then the conjunction between the sentiment word "wonderful" and the output of "very kind" is computed. Since there are no further polarity-shifting elements, the polarity of this conjunction is output as sentence polarity.

## Dependency parsing approaches

To successfully use UD-based dependency parsing for sentiment analysis as described in the previous subsection, one ideally would want the parser to be accurate, so the obtained trees provide reliable information to compute polarity, and efficient, so that the system can operate at a large scale with a reasonable resource consumption. However, choosing a parsing approach involves a tradeoff between speed and accuracy, as the most accurate systems tend to be slower (*Anderson & Gómez-Rodríguez, 2021*). In this respect, it is worth noting that *Gómez-Rodríguez, Alonso-Alonso & Vilares (2019)* showed that the polarity classification task has modest accuracy requirements, *i.e*., there is a "good enough" parsing accuracy threshold past which increasing accuracy doesn't bring further improvements to the end task. On the other hand, speed is often a practical bottleneck, especially if a large number of messages need to be processed.

However, as already noted in § Introduction, conventional syntactic parsers based on ad-hoc algorithms, like the one included in the Stanza library (*Qi et al., 2020*) are often slow and computationally costly (*Gómez-Rodríguez & Vilares, 2018*; *Roca, Vilares & Gómez-Rodríguez, 2023*; *Strzyz, Vilares & Gómez-Rodríguez, 2019b*). In recent years, however, an alternative approach has emerged that offers significantly faster parsing, based on viewing dependency parsing as a sequence labeling task. Sequence labeling can be performed in linear time with respect to the length of the input, and is highly parallelizable, so it provides high practical speeds (*Strzyz, Vilares & Gómez-Rodríguez, 2019b*; *Anderson & Gómez-Rodríguez, 2021*).

Previous work on syntactic parsing as sequence labeling has been applied to several NLP tasks such as integration of constituency and dependency parsing (*Strzyz, Vilares & Gómez-Rodríguez, 2019a*), semantic role labeling (*Wang et al., 2019*) or LLM probing (*Vilares et al., 2020*), where its speed and simplicity can be useful. CoDeLin (COnstituent and DEpendency LINearization system, *Roca, Vilares & Gómez-Rodríguez, 2023*) is a recent comprehensive library for parsing as sequence labeling that supports both constituent and dependency parsing. The basic idea behind dependency parsing in CoDeLin is to assign to each word a tag with information about a part of the dependency tree involving that word, in such a way that a full tree can be decoded from the sequence of labels for a given sentence. For this purpose, there are several types of encoding functions, each of which corresponds to a different way of representing the dependencies between words in a sentence in the form of tags associated to each of the words. We now summarize the framework and some of the encodings of interest. More information, including further encoding functions, can be found in *Strzyz, Vilares & Gómez-Rodríguez (2019b)*, *Roca, Vilares & Gómez-Rodríguez (2023)*, and *Gómez-Rodríguez, Roca & Vilares (2023)*.

Following *Roca, Vilares & Gómez-Rodríguez (2023)*, let $T_{|w|}$ be the set of dependency trees for sentences of length $|w|$. Each element of $T_{|w|}$ is a dependency tree $t_w = (V, A)$ for a sentence $w = [w_1, \ldots, w_{|w|}]$ where each $i \in V$ denotes the position of the word $w_i$, and $A$ is the set of dependency relations between words, where each $a \in A$ is of the form $a = (h, r, d)$ such that $h \in V$ is the index of the head of the relation, $d \in V$ corresponds to the index of the dependent, and $r \in R$ is a label describing a syntactic function, *i.e*., the type of syntactic relation between head and dependent (such as subject, modifier, indirect object, *etc*.). We can cast the dependency parsing task as a sequence labeling task by defining a set of labels $L$ that will let us encode each tree $t_w \in T_{|w|}$ as a sequence of $|w|$ labels $l_w \in L^{|w|}$, and an encoding function $F_{|w|} : T_{|w|} \rightarrow L^{|w|}$ that allows us to translate the trees into those labels. $F_{|w|}$ must be injective, so that each tree is encoded as a unique label sequence and correct label sequences can be decoded back into trees. While injectivity is not enough to guarantee that *incorrect* label sequences that might be output by a machine learning sequence-labeling system can also be decoded (that would require surjectivity, which known encoding functions do not have), such sequences are handled in practice with straightforward heuristics, like ignoring labels that are contradictory or violate tree constraints.

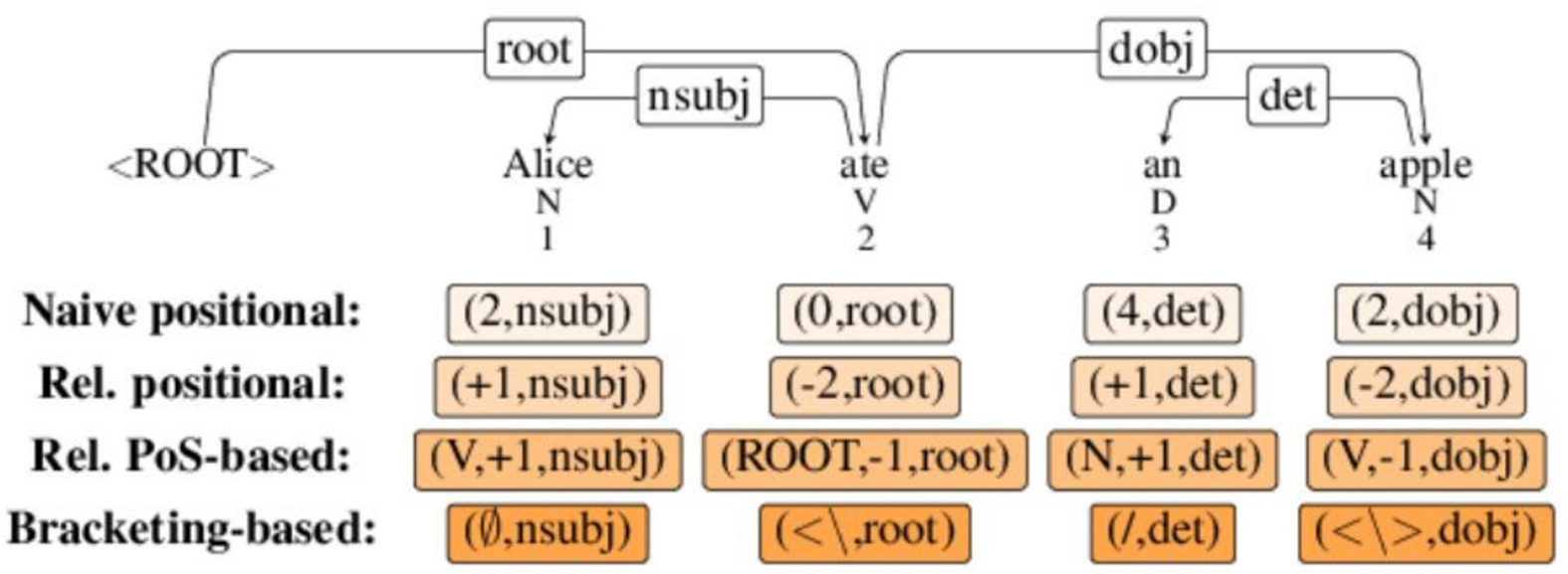

**Figure 2 A short version for demonstration of sequence labeling from *Strzyz, Vilares & Gómez-Rodríguez (2019b)*.** Full-size DOI: 10.7717/peerj-cs.3519/fig-2

In all encoding functions that we will consider here, the $j$th label in the sequence $l_w$ assigned to a sentence $w$ is a tuple $l_j = (x_j, r_j)$, where $r_j$ is always the relation type (syntactic function) describing the dependency from $w_j$'s parent to $w_j$; whereas $x_j$ is an encoding-specific value that encodes the location of the head of $w_j$.[2] In particular, we consider three encodings, where $x_j$ is defined as follows:

- **Absolute encoding**: $x_j$ directly represents the absolute position of the head of $w_j$, *i.e.*, $x_j = i \mid (i, r, j) \in A$.
- **Relative encoding**: $x_j$ represents the position of the head of $w_j$ as an offset or displacement of the head with respect to the dependent, *i.e.*, $x_j = i - j \mid (i, r, j) \in A$. Thus, a positive value $+x$ means that the head is $x$ positions to the right of $w_j$, whereas a negative value $-x$ means that it is $x$ positions to the left.
- **POS-encoding**: $x_j$ is in turn a tuple $(p_j, o_j)$, where $p_j$ is the part-of-speech tag of the head of $w_j$, and $o_j$ is a displacement where only words tagged $p_j$ are taken into account. Thus, a label $(p, +x)$ for a positive value $x$ means that the head is the $x$th among the words tagged $p$ that are to the right of $w_j$, while $(p, -x)$ means that it is the $x$th word tagged $p$ to the left of $w_j$.

Figure 2 illustrates how a sentence's parse tree is represented in terms of different encodings in CoDeLin (*Strzyz, Vilares & Gómez-Rodríguez, 2019b*; *Roca, Vilares & Gómez-Rodríguez, 2023*). Once one has chosen a tree encoding, training a parser is straightforward: a sequence-labeling model is trained to label sentences with the syntactic labels in the selected encoding, and at test time, the generated labels are decoded back to trees by using the displacements (and PoS tags, in the PoS-based encoding) to find the head of each node and build the corresponding dependency. If the model outputs a wrong label that causes a contradiction or constraint violation (*e.g*., a head offset pointing to a non-existing word position like position 6 in a sentence of length 4, or causing a cycle), that label is simply ignored and no dependency is created.

[2] In the general framework of dependency parsing as sequence labeling, labeling, $x_j$ encodes head information that does not necessarily correpond to the head of $w_j$. For example, in bracketing encodings, information about each dependency is distributed between the labels of the head word and the dependent word, rather than being encoded exclusively in the dependent's label. However, we are not using such encodings in this article. For the encodings that we use, it is true that $x_j$ encodes the location of the head of $w_j$.

**Table 1** **Polarity expression and corresponding polarity from the English training dataset in *Barnes et al. (2021)*.**

| Polarity expression | Polarity |
|---|---|
| My best honeymoon. | Positive |

Until now, parsing as SEL (as provided *e.g*., in CoDeLIn) has not been applied yet to the domain of SA using SEL-based dependency models. In this study, we fill this gap in previous research, to see if this fast parsing approach can improve the speed of syntactic rule-based sentiment analysis, and at what cost (if any) to its accuracy.

# MATERIALS AND METHODS

## Data

We use distinct datasets for each task in our pipeline. For dependency parsing, we use the datasets UD EWT English (https://github.com/UniversalDependencies/UD_English-EWT/tree/master) (*Silveira et al., 2014*) and UD Spanish AnCora (https://github.com/UniversalDependencies/UD_Spanish-AnCora/tree/master) (*Taulé, Martí & Recasens, 2008*) with standard splits.

For polarity classification, we use hotel review datasets from the OpeNER project (https://github.com/jerbarnes/semeval22_structured_sentiment) (*Agerri et al., 2013*), consisting of 1,743 reviews in English and 1,438 in Spanish.

It contains an opinion mining *corpus* of hotel reviews for six languages (de, en, es, fr, it, nl) extracted from different booking sites from November 2012 to November 2013 (*Barnes et al., 2022*). For our purposes, we only consider the English and Spanish datasets, OpeNERes and OpeNERen. Each review is annotated with individual polarity expressions and their polarity (positive or negative) as demonstrated by a simple example of a review that only contains one polarity expression in Table 1 from *Barnes et al. (2022)*. Since our problem of interest is polarity classification, we perform the following data preprocessing: ignore the information of the respective holders and targets of polarity expressions, as well as the polarity strength such as "strong" and "standard" in the dataset. We also discard 350 reviews in English and 186 reviews in Spanish that do not contain any polarity expression at all (no annotation provided in the datasets), so our evaluation is conducted on the remaining reviews that do contain subjectivity. If a review contains multiple polarity expressions—for example, "This hotel is expensive, but the staff is nice", each expression is labeled with its own polarity (*e.g*., [negative, positive]) (*Barnes et al., 2022*). Then, the aggregate polarity for the review as a whole is computed as the majority value in that list, or a third polarity (neutral) if there is no majority value (*e.g*.,[negative, positive]). We thus have a task of predicting three polarity labels (neutral, positive, and negative) on this review dataset.

Table 2 **TripAdvisor review in Spanish with corresponding polarity from the Rest-Mex dataset (*Álvarez-Carmona et al., 2023*).**

| Review | Polarity |
|---|---|
| Justo lo que buscaba. Sabores exóticos, buena atención, lugar tranquilo y bonito. Full recomendado. El sector también es tranquilo.<br>'Just what I looked for. Exotic flavors, good attention, quiet and nice location. Full recommendation. The area is also very quiet.' | 5 |

In addition, in the case of Spanish, we also use data from the Rest-Mex 2023 sentiment dataset (https://sites.google.com/cimat.mx/rest-mex2023) from TripAdvisor which contains 251,702 reviews in total, where Spanish-speaking Mexican Spanish tourists provide their evaluation or judgments on hotels, sightseeing, and restaurants (*Álvarez-Carmona et al., 2023*). An example review can be seen in Table 2. The Rest-Mex dataset contains a training dataset with polarity values accessible to the public, which was collected from various tourist destinations in Mexico, Cuba, and Colombia. The training data set includes labeled information about polarity, type of attraction, the country of origin for each opinion as well as the user-written title associated with the whole review, such as "recommendable!". This collection was obtained from tourists who shared their opinions on TripAdvisor between 2002 and 2023. Each opinion's polarity is given on a five-star scale, *i.e*., as an integer between 1 and 5, where 1: Very bad, 2: Bad, 3: Neutral, 4: Good, 5: Very good. From the training dataset, we evaluate the polarity prediction task on a subset of first 10,000 reviews, like the one shown in Table 2.[3] We do not use titles for the polarity prediction task. In order to make the accuracy results from the polarity prediction task on the Rest-Mex reviews comparable with the accuracy results performed on the OpeNER datasets, as well as to facilitate straightforward and fair comparison with our VADER baseline, we convert the 5 polarity values associated with the 10,000 reviews from the Rest-Mex data set into 3 polarity labels (positive, negative, neutral) by merging the polarities 1 and 2 under the label "negative", polarities 4 and 5 under the label "positive" and polarity 3 under the label "neutral" (see "Appendix A" for conversion rules).

[3] Note that our system has not been trained on these 10 000 reviews, as in fact it is an unsupervised rule-based system that needs no training at all. The reviews were not used in any way in the design of the rules or to tweak any system parameter. The reason for creating an evaluation dataset from the training set rather than evaluating on the test set is that the latter is not publicly available, so evaluating on it requires involving the Rest-Mex organizers. We do perform an evaluation on the test set for comparison with a Transformer-based system, as explained below, as that is a supervised system that is trained on the training set.

For the accuracy comparison with Transformer-based models in the polarity prediction task, we focus on the Rest-Mex dataset where we use the full test set containing 107,863 reviews, since it would not be adequate to evaluate the Transformer-based supervised system on inputs on which it has been trained. The gold polarity values are not accessible to the public, which is why we send the polarity values output by the systems to the Rest-Mex organizers (*Álvarez-Carmona et al., 2023*) who return the evaluation to us. In this polarity prediction task performed on the test dataset, we do not convert polarity values from 1 to 5 to three labels "positive", "negative" and "neutral", as we have no access to them, but directly evaluate accuracy on the five-star scale.

It is worth remarking that, except in the case of the Transformer-based model, polarity data from the sentiment datasets is only used for evaluation. With the aforementioned exception, the SA systems we are considering do not use a training set, as they are unsupervised with respect to polarity classification, not needing any training or undergoing any other form of tuning or tweaking with sentiment-annotated datasets. This kind of models have been shown to lag behind supervised models in accuracy when an in-domain *corpus* is available, but have more robust accuracy when it is not, a common case in practice (*Vilares, Gómez-Rodríguez & Alonso, 2017*).

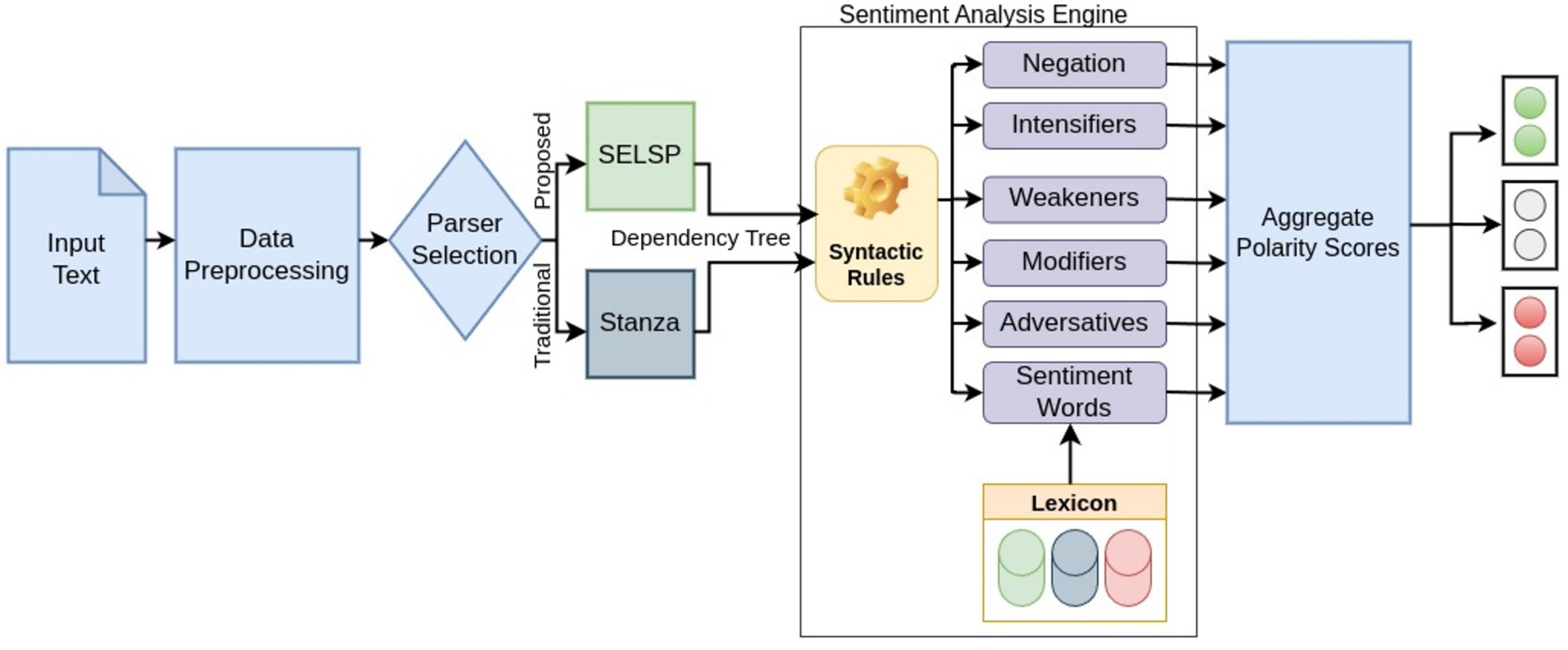

**Figure 3** **Overall architecture of the sentiment analysis system.** Full-size DOI: 10.7717/peerj-cs.3519/fig-3

It is important to note that the two datasets we are using have a significant text length difference, although the topic of the data and domain are very similar as both datasets deal with reviews of touristic destinations such as hotels and restaurants. The RestMex dataset has longer reviews than the OpeNER dataset as measured by the mean of sentences per review as well as by the mean of tokens per sentence. The mean count of sentences per review in the Rest-Mex dataset is 3.75, *vs*. 1.06 in OpeNER. The mean count of tokens per sentence in the Rest-Mex dataset is 23.77, *vs*. 16.38 in OpeNER.

The datasets and code are available in the Zenodo repository (https://doi.org/10.5281/zenodo.15323755) Moreover, the prototype (https://salsa.grupolys.org) of the system is also available.

# METHODOLOGY

The sentiment analysis methodology consists of two modules; a sequence labeling-based dependency parser (SELSP) to generate dependency trees followed by a sentiment analysis engine which uses sentiment dictionaries and syntactic rules for sentiment polarity classification from the said dependency trees as shown in Fig. 3.

**Syntactic parsing (SELSP Parser)** State-of-the-art syntactic parsers (*Kondratyuk & Straka, 2019*; *Mrini et al., 2019*) nowadays mostly use Transformer architectures due to their ability to capture complex linguistic patterns and dependencies efficiently. We specifically chose DistilBERT—a distilled version of BERT—to balance performance with computational efficiency. Unlike lightweight models (*e.g*., MobileBERT, TinyBERT), DistilBERT (*Sanh et al., 2019*) preserves 95% of BERT's performance while being 40% smaller and 60% faster. It requires less computational resources and suits well for the tasks where both accuracy and speed are crucial, such as dependency parsing. Benchmark studies (*Sanh et al., 2019*) confirm its competitive accuracy on GLUE and SQuAD, with marginal degradation compared to BERT.

To train the SELSP, we used DistilBERT-Base-Uncased as our sequence labeling classifier and evaluated the three sequence-labeling encodings mentioned in Section "Dependency parsing approaches" on the validation sets of the UD EWT English dataset (*Silveira et al., 2014*) and the UD Spanish AnCora (*Taulé, Martí & Recasens, 2008*) dataset. We chose the Relative Encoding (REL) as shown in Fig. 2, as it obtained the highest accuracy in these experiments.

We fine-tuned the DistilBERT model (*Sanh et al., 2019*) from the pretrained checkpoint (DistilBERT-base-uncased) available in the HuggingFace library using the UD-EWT English dataset and the UD Spanish AnCora dataset. DistilBERT consists of a pre-trained transformer encoder, and during fine-tuning, only the parameters of this encoder are updated. The basic idea behind fine-tuning was to leverage the knowledge learned by DistilBERT during pre-training on a large *corpus* and to adapt it for the sequence labeling task.

To fine-tune the SELSP, the original training, validation and test data splits from the considered English and Spanish UD datasets were first converted into sequence labels using the CoDeLin framework (Dependency Parsing Approaches). Then, the model was trained for 30 epochs on the resulting training set using the Hugging Face Trainer framework, with validation performed after each epoch to adjust the learned weights and biases. Training was conducted on an NVIDIA GeForce RTX 3090 GPU with a fixed random seed (seed = 42) for reproducibility. Table 3 describes the other hyperparameters used to train the model. The same hyperparameters were used to train both English and Spanish models. The pipeline of SELSP parser is given in Fig. 4.

The training process results into a model that, given an input sentence, predicts the labels encoding the dependency heads and relations in our chosen encoding, in this case the Relative Encoding (REL). We then use the CoDeLin sequence labeling framework to decode these rules produced by the SELSP into the CoNLL-U format that is standard for Universal Dependencies. With this, the accuracy of the parse trees produced by the fine-tuned SELSP was evaluated using the CoNLL 2018 standard evaluation script (*Zeman et al., 2018*).

The CoNLL-UD format represents a dependency graph in a particular format according to which tokens are annotated linguistically concerning their lemma, POS, morphology, the location of their syntactic head expressed as an absolute token index, and the label of the dependency relation such as subject relation (nsubj), object relation (obj), determiner (det), adverbial modification (advmod), conjunction (cc), conjoined element (conj), complementizer (xcomp), or punctuation (punct), as shown in Fig. 1. In this format, word tokens are indexed with integers starting from 1, and all the information about a given token is in a separate line. The information of each dependency relation is included in the line associated with the dependent, by specifying the index of its head together with the relation type. For instance, in the example in the figure, the head of the word token with the integer 2 is the word 3 (the main verb), with relation type "nsubj" indicating that "I" is the subject of "found"; and the modifier "very" at position 10 is modifying the word "kind" at position 11. The sentence's syntactic root (the main verb "found" in the example) has no head, so its associated head index is set to zero. This information generated by the dependency parser is used to identify polarity shifters: for example, since "very" is a modifier of "kind", it will increase the positive polarity associated with "kind". The polarity prediction demonstrated between the dependency graph and the CoNLL-UD format in Fig. 1 will be explained below.

Table 3 Hyperparameters to fine-tune the SELSP.

| Hyperparameter | Value |
|---|---|
| Epochs | 30 |
| Batch size | 128 |
| Learning rate | 1e–4 |
| Weight decay | 0.001 |
| Adam epsilon | 1e–7 |
| Adam beta1 | 0.9 |
| Adam beta2 | 0.999 |
| Metric for best model | f1 |
| Evaluation strategy | Epoch |

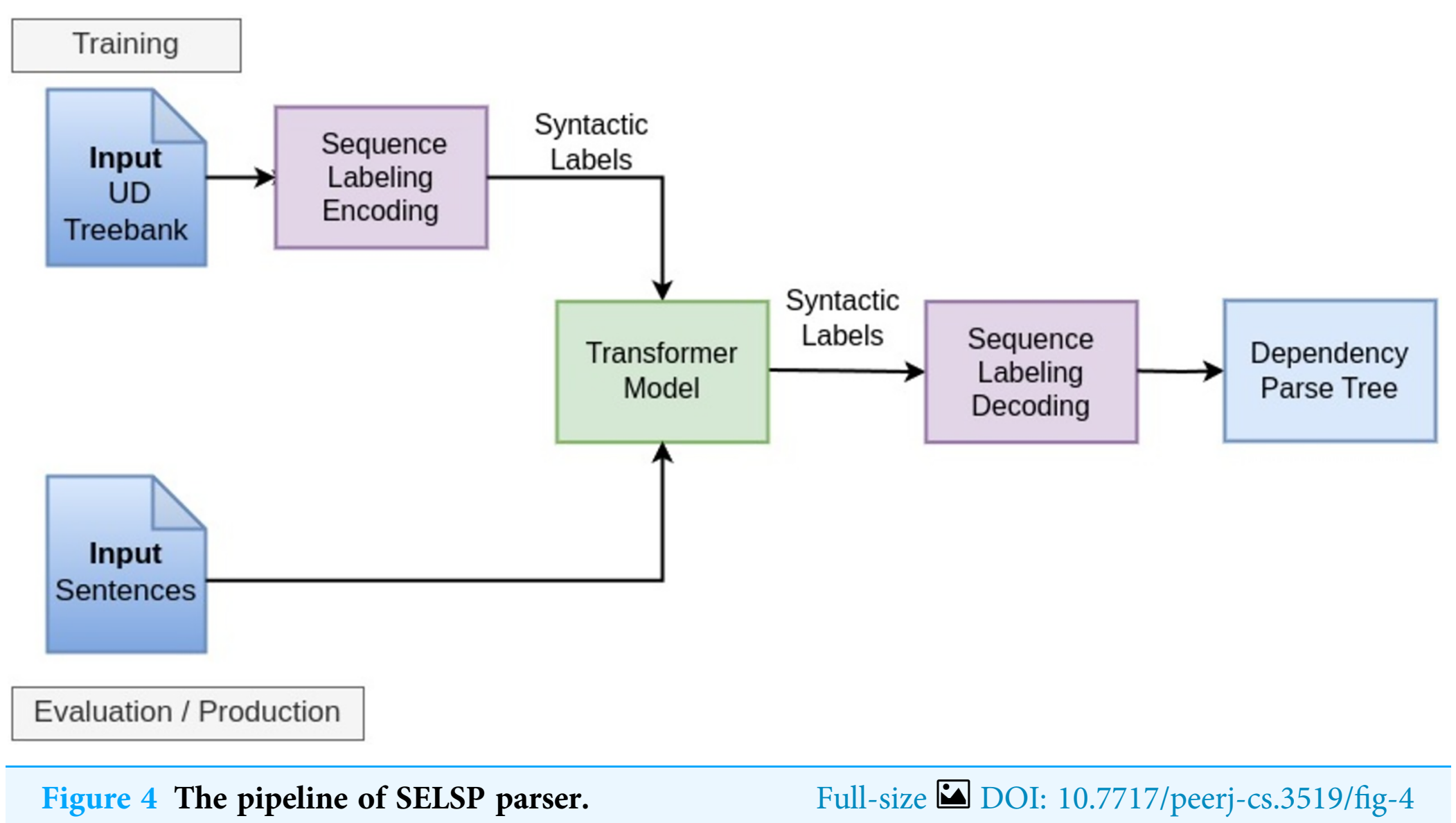

Figure 4 The pipeline of SELSP parser. Full-size DOI: 10.7717/peerj-cs.3519/fig-4

**Sentiment analysis engine for polarity classification** Once the SELSP is trained, its output (dependency tree) is fed into the sentiment analysis engine which uses syntactic rules (*Kellert et al., 2023*) to calculate sentiment polarity from said dependency trees. The overall architecture of the sentiment analysis system is provided in Fig. 3.

The original sentiment analysis system of *Kellert et al. (2023)* contains information on how to calculate polarity scores of reviews based on sentiment words from the sentiment dictionary SO-CAL (*Brooke, Tofiloski & Taboada, 2009*) and UD-based syntactic rules for Spanish, using syntactic parse trees obtained from Stanza (*Kellert et al., 2023*). Apart from incorporating the option of parsing with SELSP as described above, we also extended the system to handle both English and Spanish, as well as to be able to incorporate different sentiment dictionaries apart from SO-CAL. We now describe the extensions made.

We use a set of English negation words and intensifiers like "very" or weakeners like "few" to process English sentiment data, which can be found in "Appendix B". The set of negation words we use for processing Spanish data is integrated into the code from *Kellert et al. (2023)*.

We use several English and Spanish sentiment dictionaries. For the English sentiment dataset, we use the following dictionaries: (1) SO-CAL, which is available in Spanish and English (*Brooke, Tofiloski & Taboada, 2009*), (2) VADER Lexicon (VADERdict), which is only available in English (*Hutto & Gilbert, 2014*), (3) an aggregate version that we obtain by adding to SO-CAL the sentiment words from VADERdict that are not initially covered by SO-CAL, and (4) the English dictionary from BabelSenticNet (*Vilares et al., 2018*). For the two Spanish sentiment datasets (OpeNER reviews and the 10,000 Spanish TripAdvisor reviews from RestMex), we use (1) the Spanish version of SO-CAL, (2) an aggregate version that combines SO-CAL with words from VADERdict that are missing in the English SO-CAL that we translate into Spanish *via* DeepL (SO-CAL + sp. VADERdict), and two automatically-extracted datasets: (3) words with low ambiguity extracted from titles from the Rest-Mex training dataset (*Kellert, Gómez-Rodríguez & Uz Zaman, 2024*) and (4) an aggregate version that combines SO-CAL with words from VADERdict and Rest-Mex dictionary.

We now provide some details on how these two latter dictionaries have been generated. We measure the polarity ambiguity of words by the standard deviation of polarity judgments (sdv) (*Kellert, Gómez-Rodríguez & Uz Zaman, 2024*). The authors of said article motivate their method of constructing sentiment dictionaries by using words with low sdv from titles as follows. Titles of reviews represent short summaries of long reviews and the words from these summaries often function like sentiment words, *e.g.*, "Recommendable!", "Excellent!", *etc*. However, sometimes titles also represent topics of reviews, like "Last holidays", and the words from these topic-titles cannot be considered as sentiment words. In this case, as the authors show (*Kellert, Gómez-Rodríguez & Uz Zaman, 2024*), words from these titles tend to be associated with a high sdv, since people will provide different, sometimes opposite judgments, to reviews containing that kind of topic words. For this reason, the authors consider only words with low sdv as good candidates for building sentiment words out of titles. Thus, we only select words with lower polarity variation measured by sdv (<0.6). For the experiments in this study, we use two different word lists. One word list contains words with low sdv from short titles (ShortT), which are those that contain only one word such as "Excellent!", and the other list contains words with low sdv from short and long titles (ShortT + LongT), where long titles are those with multiple words. It is important to note that the accuracy calculation of the polarity prediction task based on the Rest-Mex dataset was performed on whole review texts excluding their titles. As a consequence, the sentiment dictionary created from the titles of the Rest-Mex dataset does not influence the accuracy of the polarity prediction task based on reviews from the Rest-Mex dataset.

Regarding aggregation of sentiment dictionaries (like when we combine SO-CAL with VADERdict), the issue can arise that two dictionaries have different polarities for the same word. Throughout the article, we will consider that SO-CAL is the basic dictionary to which we add words with corresponding polarities from other dictionaries that are not contained in SO-CAL. Thus, if SO-CAL has a lemma/word with a given polarity, but we combine it with another dictionary (such as VADERdict or the dictionaries extracted from Rest-Mex data) where the same lemma/word has a different polarity, we always keep the score assigned by SO-CAL.

For all these sentiment dictionaries, we compare the polarity results produced by our system that combines SELSP and the sentiment analysis engine with two main baselines. The first one is the version of the same SA rules that used Stanza-based syntactic parsing as their backbone instead (*Kellert et al., 2023*). The second main baseline is the shallow heuristics-based approach VADER, which is a widely used in practical applications of SA in research and industry use cases (see *e.g.*, *Lasser et al., 2023*; *Deas et al., 2023*; *Yuliansyah et al., 2024*; *Barik & Misra, 2024*; *Lahande, Lahande & Kaveri, 2025* for some recent examples).

We also compare our results from the Spanish polarity prediction task with the state-of-the-art Transformer-based model from the javier-alonso team that achieved the second rank in the polarity prediction task in the Rest-Mex shared task 2023 (*Álvarez-Carmona et al., 2023*; *Alonso-Mencía, 2023*). The javier-alonso team (*Alonso-Mencía, 2023*) used two variants of the RoBERTa model, namely roberta-base-bne and twitter-xlm-roberta-base. They balanced the effect of minority classes (such as polarity 1 and 2) through oversampling strategies. The reason why we used the Transformer-based model from the second highest ranked team and not the first is that their source code was easily accessible to us. Note, however, that the results from both teams in the shared task were very similar, with a difference of only one percentage point in the polarity classification task. Thus, we can assume that the results of this system are a reasonable representation of the state of the art.

For the comparison with the Transformer-based model, we use the following dictionaries: (1) words with low ambiguity extracted from short and long titles from the Rest-Mex training data as explained before (*Kellert, Gómez-Rodríguez & Uz Zaman, 2024*) and (2) a combination of dictionaries: the Spanish version of SO-CAL, together with words from the VADER dictionary that are missing in English SO-CAL that we translate into Spanish *via* DeepL (sp. VADERdict), and words with low ambiguity extracted from short and long titles from the Rest-Mex training data (*Kellert, Gómez-Rodríguez & Uz Zaman, 2024*).

Table 4 **Comparison between SELSP and Stanza on English and Spanish universal dependencies treebanks.** Best results for each language are highlighted in bold.

| Treebank | Parser | Adverb modifier F1 | Adjective modifier F1 | Subject F1 | Object F1 | UAS F1 | LAS F1 | MLAS | BLEX |
|---|---|---|---|---|---|---|---|---|---|
| UD EWT English | SELSP | **75.86** | 73.79 | **97.14** | **80.37** | 79.43 | 76.96 | **72.19** | 74.14 |
| UD EWT English | Stanza | 70.30 | **75.56** | 77.33 | 77.01 | **86.17** | **85.82** | 8.71 | **82.39** |
| UD Spanish AnCora | SELSP | **86.99** | 90.05 | **94.07** | **92.84** | 86.67 | 77.84 | **69.29** | 71.24 |
| UD Spanish AnCora | Stanza | 81.24 | **92.09** | 91.62 | 83.85 | **89.62** | **89.10** | 16.26 | **79.56** |

One of the reasons why we experiment with several dictionaries and their combinations is to see which dictionary or combination produces better accuracy results in polarity prediction tasks, both within the same system and across different systems.

For the evaluation of the systems on the polarity prediction task, we converted the outputs of these systems into class labels for ternary classification ([positive, neutral, negative]) (see "Appendix A" on conversion rules), which we then compared with the gold polarity classes obtained from the datasets in the same format (Data). As an exception, for the comparison with the Transformer-based model, we perform the evaluation directly on the five-star scale (*i.e.*, 5-class classification) due to the restrictions inherent to this evaluation: as mentioned before, the model has been trained and adjusted on the training set, so we must use the test set to which we do not have direct access, performing instead the evaluation by sending system outputs to the task organizers. Therefore, no conversion is possible.

## RESULTS AND DISCUSSION

We first report our results for dependency parsing, which serves as the basis for the subsequent polarity classification. The parsing results in Table 4 include both overall parsing metrics (UAS, LAS, MLAS and BLEX) and fine-grained metrics for some specific dependencies that are especially important for sentiment analysis.[4] In the overall metrics, our SELSP falls short of state-of-the-art accuracy for the UD EWT English and UD Spanish AnCora treebanks, and is also less accurate than Stanza, achieving a few points over 75% LAS. However, if we look at the detailed results, the picture is different. In spite of Stanza's higher LAS, SELSP is highly effective on specific dependency types that are crucial for sentiment analysis. For example, it is more accurate at detecting adverbial modifiers (advmod), subjects (nsubj), and objects (obj) across both English and Spanish. These dependencies are vital to capture intensifiers (*e.g.*, "very good") and connect the sentiment to the subjects and objects involved (*e.g.*, "I love this book").

In this respect, it is worth reminding that syntax-based polarity classification does not need high parsing accuracy (*Gómez-Rodríguez, Alonso-Alonso & Vilares, 2019*) so our goal here is to optimize for speed (hence using a sequence-labeling approach and a small distilled language model, decisions that sacrifice accuracy in favor of efficiency). This goal is evaluated in Table 5, where we report the speed of the whole system (including dependency parsing and polarity classification). Indeed, our results show that our SELSP-based SA system is much less time-consuming than the alternative using the conventional syntactic parser Stanza as well as the VADER system, as it processes almost three times more sentences per second than Stanza and 18 times more than VADER. Speed does not vary substantially with respect to the sentiment dictionary used, with only a slight slowdown when using larger dictionaries. SELSP's speed advantage is attributed to its sequence-labeling formulation, which contrasts with the computationally intensive *ad hoc* parsing algorithms employed by Stanza. By reformulating parsing as a linear-time sequence-labeling task (see "Dependency Parsing Approaches"), SELSP achieves very fast processing speeds.

[4] UAS (Unlabeled Attachment Score) measures the percentage of words that are assigned the correct syntactic head, regardless of the dependency label. LAS (Labeled Attachment Score) additionally requires the dependency label to be correct. MLAS (Morphology-Aware LAS) extends LAS by also evaluating morphological features. BLEX (Bi-Lexical Dependency Score) is similar to MLAS but instead of morphological aspects, it includes lemmatization in the evaluation. As fine-grained metrics, we show F1-scores for specific dependency types that are important for sentiment analysis.

**Table 5** **Comparison between SELSP, Stanza, RoBERTa and VADER in processing speed (instances, sentences, tokens/second).** For the speed, the systems are run on a GPU: NVIDIA GeForce RTX 3090. Best results are highlighted in bold.

| System | Dictionary | Instances/Sec. | Sentences/Sec. | Tokens/Sec. |
|---|---|---|---|---|
| SELSP | SO-CAL | 48.30 | 181.78 | 3,380.72 |
| SELSP | SO-CAL + VADER | 49.12 | 184.85 | 3,437.95 |
| SELSP | Rest-Mex | **54.15** | **203.81** | **3,790.41** |
| SELSP | Rest-Mex + SO-CAL + VADER | 50.64 | 190.60 | 3,544.80 |
| Stanza | SO-CAL | 14.51 | 54.61 | 1,015.62 |
| Stanza | SO-CAL + VADER | 14.14 | 53.23 | 990.00 |
| Stanza | Rest-Mex | 14.64 | 55.10 | 1,024.76 |
| Stanza | Rest-Mex + SO-CAL + VADER | 13.70 | 51.56 | 958.91 |
| VADER | | 3.74 | 14.06 | 261.53 |
| RoBERTa | | 29.34 | 110.43 | 2,053.71 |

**Table 6** **Comparison between SELSP, Stanza and VADER in accuracy (percent) in the polarity prediction task performed on the English dataset from OpeNER across several dictionaries (see Section "Methodology") based on three polarity labels "neutral", "positive" and "negative".** Best results for each language are highlighted in bold.

| Dictionary | SELSP | Stanza | VADER |
|---|---|---|---|
| SO-CAL | 78.89 | **79.61** | |
| VADERdict | 76.16 | **76.45** | 69.92 |
| SO-CAL + VADERdict | **78.17** | 77.49 | |
| BabelSenticNet | **76.38** | 76.09 | |

**Table 7** **Comparison between SELSP, Stanza and VADER in accuracy (percent) in the polarity prediction task performed on the Spanish dataset from OpeNER across several dictionaries based on three polarity labels "neutral", "positive" and "negative".** Best results for each language are highlighted in bold.

| Dictionary | SELSP | Stanza | VADER |
|---|---|---|---|
| SO-CAL | **80.51** | 79.47 | |
| sp.VADERdict | | | 72.44 |
| SO-CAL + sp.VADERdict | **80.59** | 78.83 | |
| Rest-Mex | 80.27 | **80.35** | |
| SO-CAL, VADERdict, Rest-Mex | **78.03** | 77.55 | |

**Table 8 Comparison between SELSP, Stanza and VADER in accuracy (percent) in the polarity prediction task in the Spanish Rest-Mex training dataset (10,000 reviews) across dictionaries based on three polarity labels "neutral", "positive" and "negative".** Best results for each language are highlighted in bold.

| Dictionary | SELSP | Stanza | VADER |
| --- | --- | --- | --- |
| SO-CAL | **91.86** | 90.26 | |
| VADERdict | | | 91.23 |
| SO-CAL + sp. VADERdict | **92.03** | 90.78 | |
| Rest-Mex | **93.77** | 92.61 | |
| SO-CAL, VADERdict, Rest-Mex | **94.86** | 94.32 | |

Tables 6, 7, and 8 show the accuracy results for SELSP, Stanza, and VADER for different datasets and different dictionaries (see "Methodology"). Compared to the baselines, the results demonstrate that SELSP achieves significantly better performance than VADER across all language, dataset, and dictionary combinations. This improvement stems from fundamental methodological differences: while VADER relies on shallow heuristics without syntactic analysis, SELSP explicitly models syntactic relationships (*e.g.*, negation scope, intensifier modification) through dependency parsing, enabling more precise polarity predictions, particularly in complex sentences. The accuracy gain over VADER highlights the importance of syntactic modeling for polarity shifts. When comparing SELSP to Stanza, we observe that the accuracy of the system using the SELSP parser is not only on par with the version using Stanza, but even outperforms it in most cases (nine dataset-dictionary combinations where SELSP provides better accuracy than Stanza, *vs.* three where Stanza is better). This is particularly interesting, given that our goal in replacing Stanza with SELSP was to optimize for efficiency, and the SELSP parser prioritizes speed over accuracy (Table 4) and in particular is less accurate than Stanza in overall terms. The fact that this does not result on any loss of accuracy on the end task is a confirmation of the previous observation by *Gómez-Rodríguez, Alonso-Alonso & Vilares (2019)* that syntax-based polarity classification performance no longer improves with parsing accuracy once a "good enough" parsing accuracy threshold is achieved. Clearly, the accuracies provided by SELSP in Table 4 are above that threshold for our datasets under consideration, especially when considering that accuracies are especially good on dependencies that are crucial for sentiment analysis, so our results confirm that it is beneficial to use parsing as sequence labeling in this scenario.

If we compare dictionaries, we observe that SO-CAL provides the best results for English (and augmentation with VADERdict does not help), although differences are not large. On Spanish, there are no large differences between the different dictionaries we consider, except that the sentiment dictionaries automatically generated from Rest-Mex titles provide better results on the Rest-Mex dataset, suggesting that it is beneficial to generate dictionaries from in-domain data when possible.

**Table 9 Comparison between SELSP, Stanza and Transformer-based model implemented with RoBERTa in accuracy (percent) in the polarity prediction task in the Spanish Rest-Mex test dataset (reviews only) across dictionaries based on five polarity values.** Best results for each language are highlighted in bold.

| Dictionary | RoBERTa | SELSP | Stanza |
| --- | --- | --- | --- |
| SO-CAL | | **42.81** | 41.23 |
| SO-CAL, VADERdict | | **40.37** | 38.97 |
| Rest-Mex | | 59.55 | **60.10** |
| SO-CAL, VADERdict, Rest-Mex | | 57.78 | **58.36** |
| No Dict | **68.75** | | |

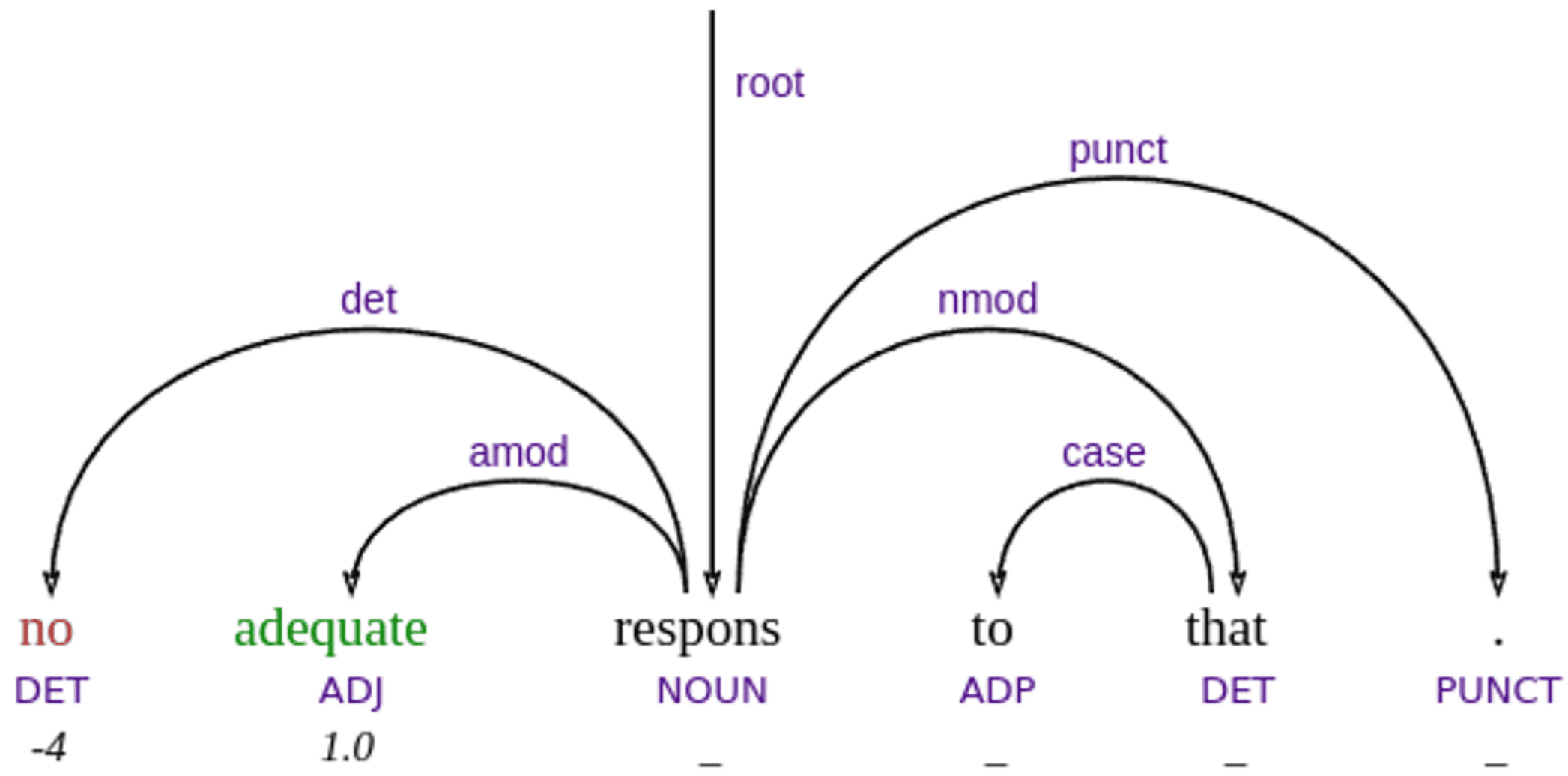

**Figure 5** Example of negation handling. Full-size DOI: 10.7717/peerj-cs.3519/fig-5

In terms of datasets, our sample of Rest-Mex reviews are clearly easier to classify than the OpeNER reviews, as every system provides better accuracy on Rest-Mex than on the other datasets.

Finally, we compare our models to a supervised Transformer-based model on the Rest-Mex Spanish dataset. Results are shown in Table 9. Note that, as explained in Section "Methodology", this comparison is performed on a different split (the Rest-Mex test set) and we perform five-class classification, rather than three class, so results are not directly comparable to those in Table 8. The table shows a clear advantage of the Transformer-based model in terms of accuracy. This result is not surprising, since supervised models are known to excel when trained on a *corpus* with similar texts as the test data (*Vilares, Gómez-Rodríguez & Alonso, 2017*), as is the case here. The use case for unsupervised lexicon-based models, like the rest of the systems considered in this article, is when no in-distribution data is available to train a supervised model, a common situation in production.

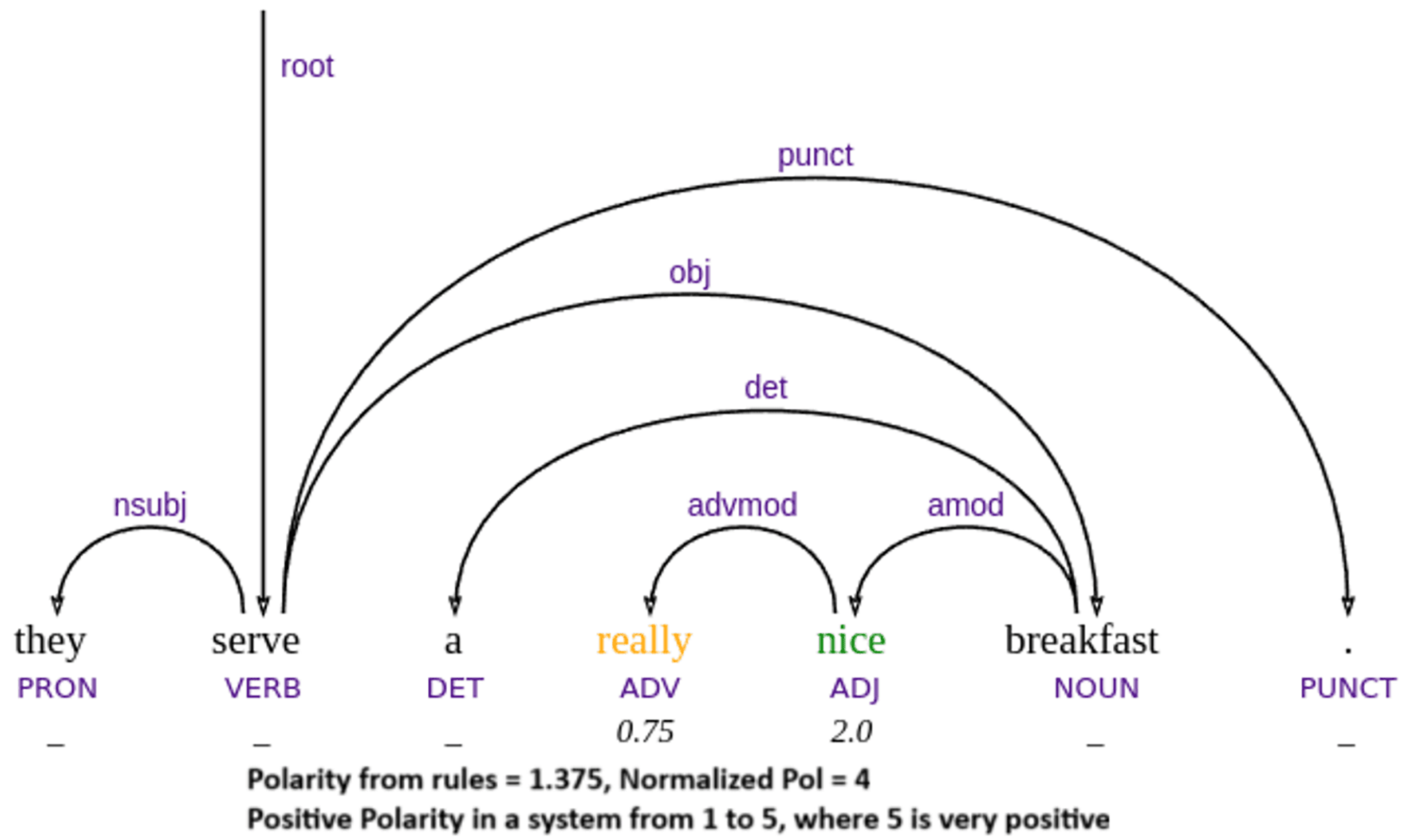

**Figure 6 Example of adverbial modifier handling.** Full-size DOI: 10.7717/peerj-cs.3519/fig-6

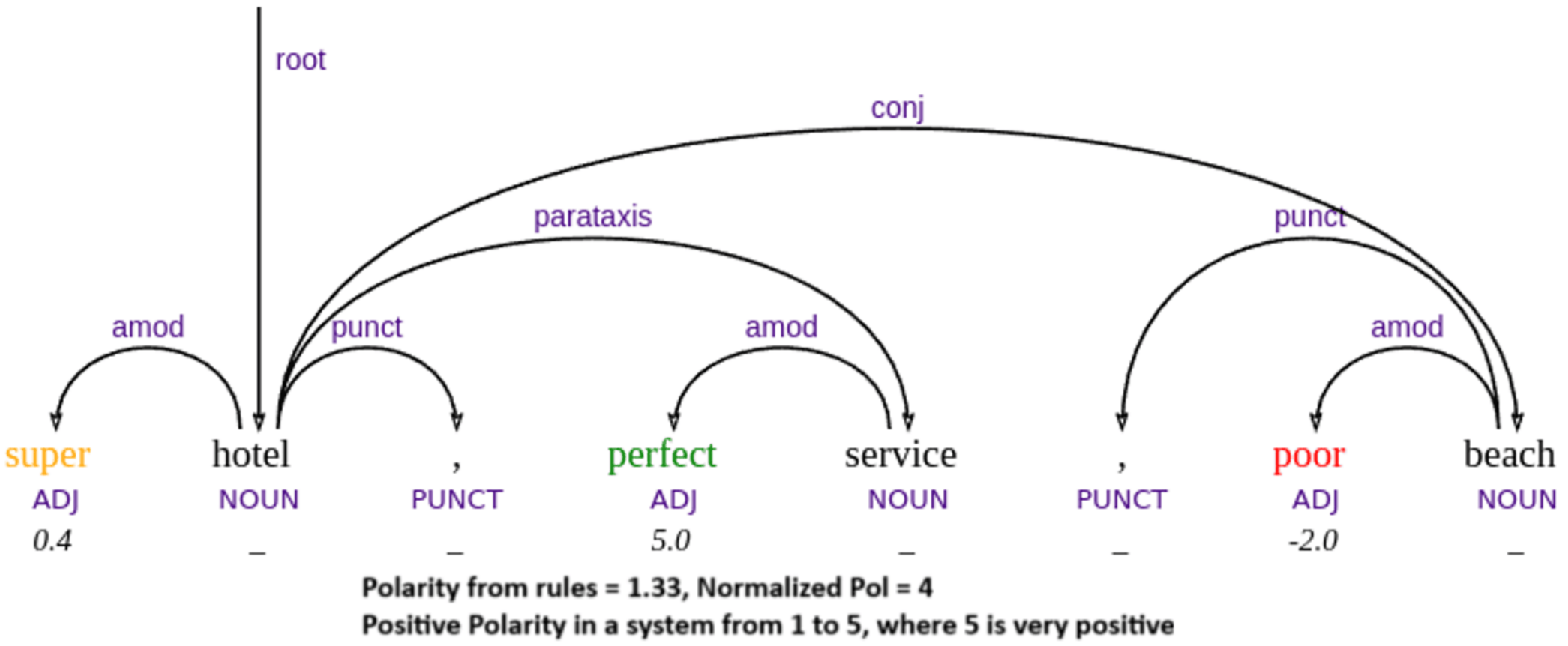

**Figure 7 Example of handling mixed polarities.** Full-size DOI: 10.7717/peerj-cs.3519/fig-7

To better illustrate the strengths and limitations of our approach, we now present a qualitative analysis of specific representative examples from the test data, including success and failure cases. Our parser's performance demonstrates that it achieves its best results in compositional cases, where the combination of word meaning and syntactic structure naturally leads to the correct sentiment outcome. In such instances, the parser effectively applies syntactic rules to interpret sentiment polarity, as illustrated in example Figs. 5, 6 and 7. For example, when a sentiment-bearing word is negated, the parser accurately reverses the polarity (*e.g.*, recognizing that "no adequate" conveys a negative sentiment in example Fig. 5). These cases highlight the strength of the system in handling syntactically transparent constructions, where sentiment can be reliably derived from grammatical composition rather than from contextual or idiomatic interpretation.

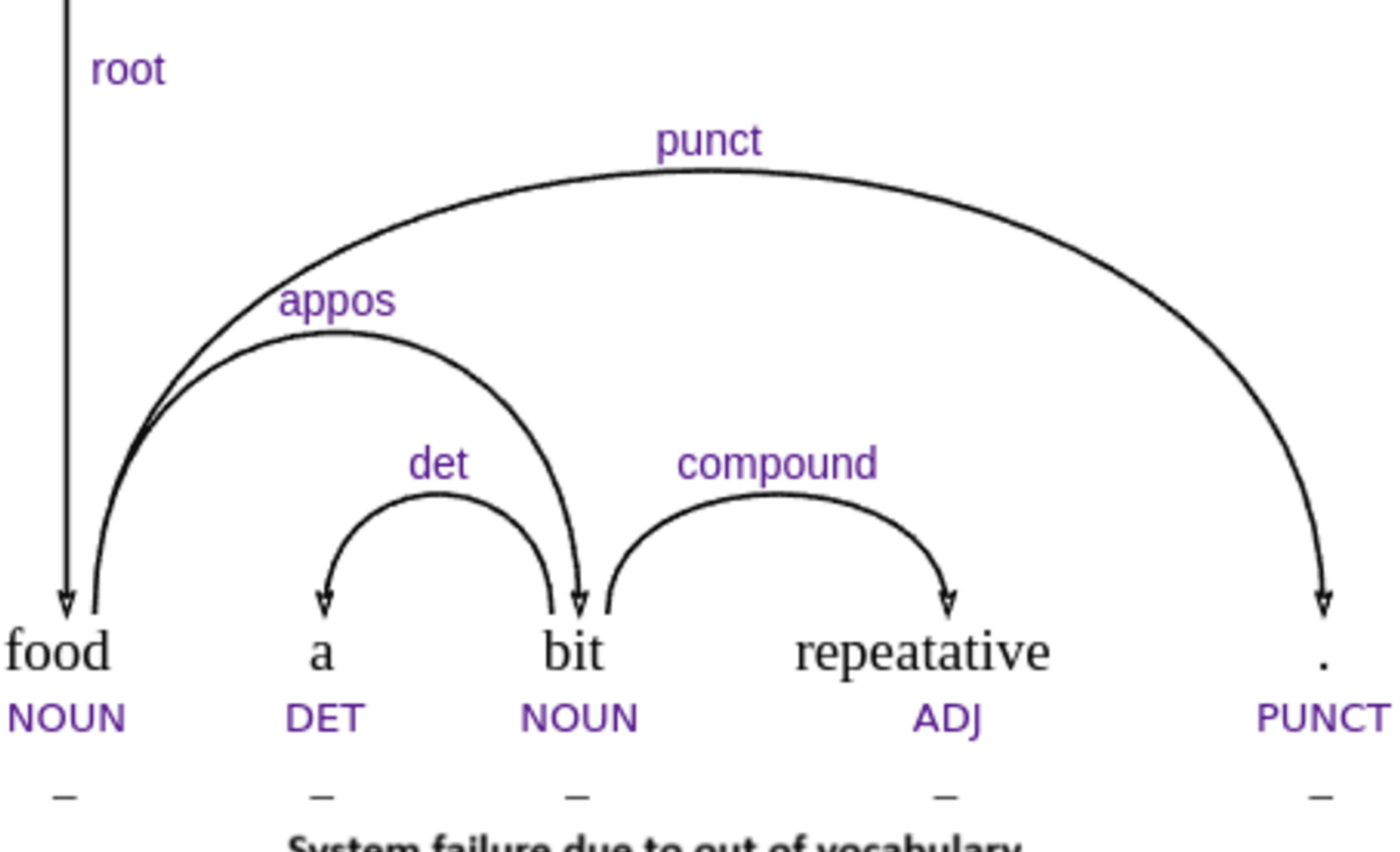

**Figure 8** Example of system failure due to out of vocabulary word (caused by a typo).
Full-size DOI: 10.7717/peerj-cs.3519/fig-8

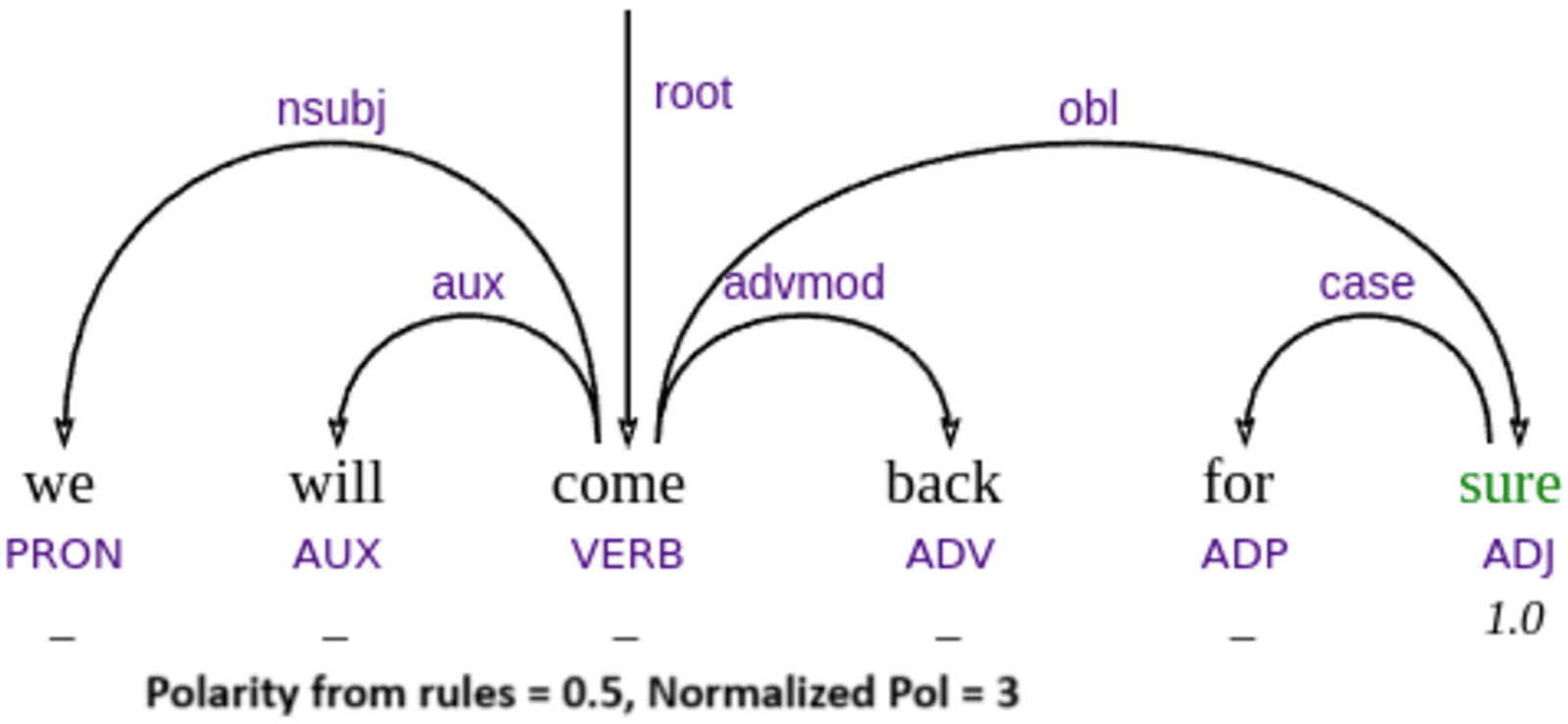

**Figure 9** Example of misclassification due to words with low polarity value.
Full-size DOI: 10.7717/peerj-cs.3519/fig-9

Our parser's performance is not independent of the sentiment dictionaries it relies on. Therefore, prediction errors are not necessarily due to failures in parsing but often result from missing or incomplete lexical entries, as illustrated by example Fig. 8. However, we note that example Fig. 9 is not strictly out-of-vocabulary (OOV), since the word "sure" is in the vocabulary. The issue lies in the low positive polarity score, which is insufficient to trigger a positive classification, leading the system to output a neutral sentiment.

Another limitation is that our approach currently associates each word with a single sentiment value, without considering contextual information or ambiguity. For instance, the word taste may be interpreted as positive in "this food has taste", but negative in "this has a perfume taste", where it conveys an artificial or unpleasant flavor—as seen in one of our failure cases in Fig. 10. In example Fig. 9, "we will come back" also involves implicit sentiment, since the surface meaning simply describes an action, but pragmatically it implies a positive evaluation—if one returns, it is because the experience was favorable.

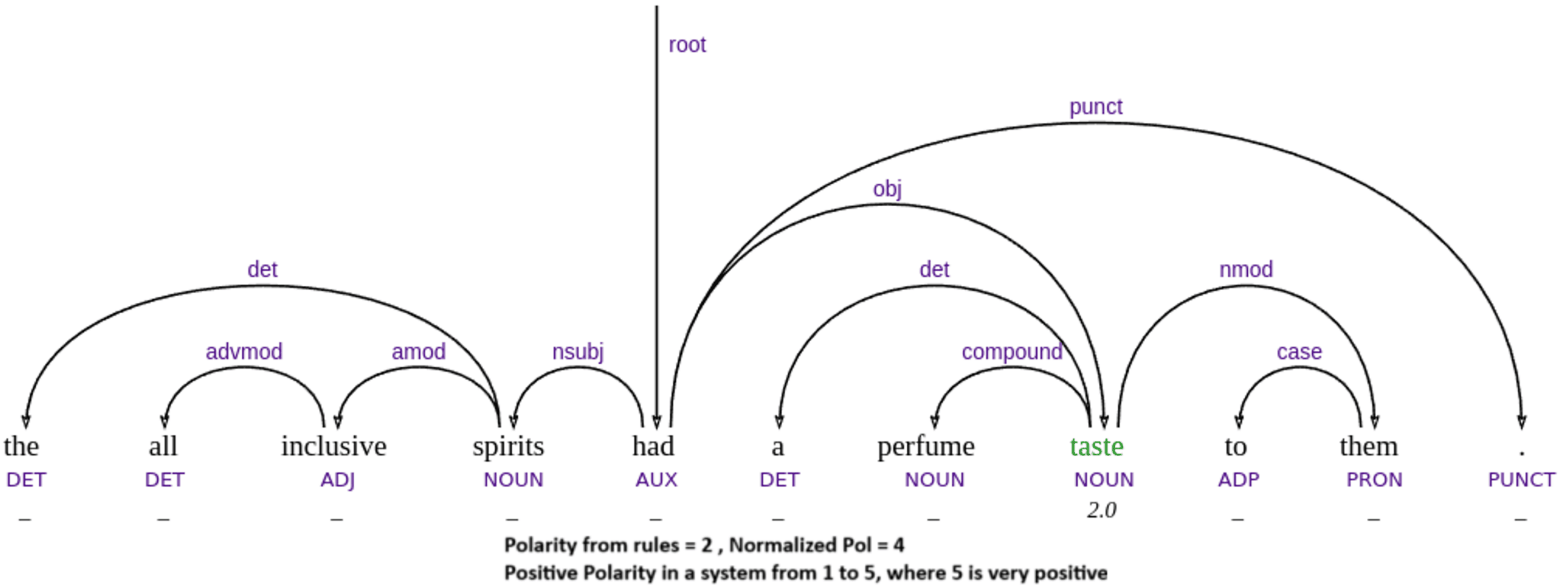

**Figure 10** **Example of failure to deal with implicit sentiment.** Full-size DOI: 10.7717/peerj-cs.3519/fig-10

Furthermore, misclassifications can occur when sentiment-bearing words are written in nonstandard forms or contain spelling errors, preventing proper dictionary matching (see example Fig. 8). Overall, sentiment analysis errors are not purely syntactic in nature nor solely dependent on the parser's ability to identify sentiment structure; they also depend on external factors such as dictionary coverage, lemma identification, and the system's sensitivity to context.

## CONCLUSION

In this study, we have addressed the speed bottleneck problem of sentiment analysis systems that use syntactic parsing. For this purpose, we trained a model for syntactic parsing as sequence labeling (SELSP) based on a pre-trained language model DistilBERT within the CoDeLIn formalism for sequence labeling. We then applied it to sentiment analysis on Spanish and English datasets, by using its output parse trees to compute sentence polarities from sentiment words through a sentiment analysis engine which relies on syntax-based rules.

Our main experiments on 3-class polarity classification on a range of English and Spanish datasets, and with a variety of sentiment dictionaries, show that SELSP achieves considerably faster processing time than the same syntax-based rules applied together with the conventional syntactic parser Stanza, as well as the popular heuristic approach VADER. In terms of accuracy, SELSP achieves better results than VADER across all datasets and sentiment dictionaries tested, and it even outperforms the Stanza-based alternative in most dataset-dictionary combinations. This shows that sequence-labeling parsing is especially fitting for syntax rule-based sentiment analysis, as we achieved substantial speed gains with no cost (or even with improvements) to accuracy. While this work focuses on sentiment analysis, the SELSP framework could potentially benefit other NLP tasks reliant on syntactic parsing, such as semantic role labeling or relation extraction. Investigating these applications remains an avenue for future research.

To put our results into context, we also ran a comparison with a supervised system based on variants of the Transformer-based model RoBERTa, which achieved state-of-the-art results at the Rest-Mex 2023 shared task. The supervised system achieved worse speed, but better accuracy, than our proposal in our comparison. This accuracy advantage is expected, as supervised systems trained or fine-tuned with in-distribution data (in this case, with more than 250K instances of reviews of the same type it is being evaluated with) tend to outperform unsupervised, rule-based systems as they can adapt specifically to the characteristics of the training data. However, this trend has been shown to often invert when one does not have in-domain training data, or even when the data is in-domain but from a different source or time period (see *Vilares, Gómez-Rodríguez & Alonso, 2017*). For this reason, unsupervised approaches like the ones we focus on here are still popular in industry applications, where it is often not possible (or very costly) to obtain large amount of data that closely resembles production data.

The improved processing times and the improved accuracy performance in 3-class categorizations (neutral, positive, and negative), together with not needing any training data and the explainability afforded by the parse trees and syntactic rules, make SELSP very attractive for practitioners of Sentiment Analysis in research and industry.

We noticed a big difference in accuracy across datasets (OpeNER *vs*. TripAdvisor reviews). This difference is not specific to our proposed system, but also common to all baselines, and it may be related with the length of the reviews (see "Data"). We believe that the performance of SELSP in processing short reviews could be improved in the future by using specialized large language models that were pre-trained on short messages like tweets, such as BERTweet (*Nguyen, Vu & Nguyen, 2020*). As our model depends on the syntactic relation between sentiment words and polarity shifters like negation, improving these rules can also lead to an improvement in accuracy.

We also tested different sentiment dictionaries, to test their influence on polarity classification results as well as their interaction with different systems. In this respect, we observed that accuracy values differ more across models (SELSP, Stanza and VADER) than across dictionaries of the same model in OpeNER. For instance, the differences between dictionaries within the SELSP model in processing English reviews are in the 3% range, whereas the differences in accuracy between models such as SELSP and VADER are in the 10% range. Similar observations can be made for the Spanish reviews from OpeNER. This suggests that the choice of a model or approach of syntactic representation is more important for the accuracy of polarity prediction tasks than the choice of a dictionary, at least for processing reviews from OpeNER. However, if we consider the longer reviews from the Rest-Mex dataset, differences between dictionaries are comparable to those between models, as shown in Table 8.

# LIMITATIONS

In our experiments, we do not perform any optimization or other task-specific tweaks to try to improve the accuracy of the polarity prediction task. This is because our main focus in this article is not to improve the accuracy of the Sentiment Analysis but rather to improve the temporal restrictions of syntax-based sentiment analysis systems to leverage their advantages such as explainability and transparency in practical settings, in contrast to purely supervised approaches. We note that, even under such a setup, the accuracy of our model is good, even outperforming the Stanza-based model apart from the VADER baseline. Another limitation of our approach is that we only use only English and Spanish to train our model (Task 1) and to measure the accuracy of the polarity prediction task (Task 2). This is because our approach requires language-specific sentiment dictionaries and language-specific words to identify polarity-shifting elements like negation and intensification. We will acquire these resources for other languages in future research to be able to include more languages for testing and evaluation. While the result that parsing accuracy beyond a "good enough" threshold is enough to help parsing accuracy was originally only shown for English (*Gómez-Rodríguez, Alonso-Alonso & Vilares, 2019*), our results suggest a similar trend for Spanish, making us expect that it will be generalizable to other languages, although further research is needed. Finally, we acknowledge that converting sentiment polarities to a ternary scale (Appendix A) could introduce some noise, but such conversions are unavoidable when testing on diverse datasets.

# APPENDIX A

## Conversion rules

**From SO-CAL polarity to 1–5 Polarity:**

from −5 to −3 = 1
from −3 to −1 = 2
from −1 to +1 = 3
from +1 to +3 = 4
from +3 to +5 = 5

**Normalize Sentiment Score to 1–5 Polarity:**

$$Polarity = \left(\frac{\text{value} - \text{min_in}}{\text{max_in} - \text{min_in}}\right) \cdot (\text{max_out} - \text{min_out}) + \text{min_out}$$

**From 1–5 Polarity to Polarity Labels:**

from 0 to 2 = "Negative"
from 2 to 3 = "Neutral"
from 3 to 5 = "Positive"

**From VADER scores to Polarity Labels:**

compound score ≤ −0.05 = "Negative"
−0.05 < compound score < 0.05 = "Neutral"
compound score ≥ 0.05 = "Positive"

# APPENDIX B

## Lists of English negation words and modifiers

**List of modifiers:** fugging, so, majorly, blisteringly, consistent, phenomenally, fewer, complete, frackin, abundantly, utmost, definite, awfully, hugely, bigger, insignificant, barely, deeply, somewhat, moderate, radically, quite, arguably, little, fracking, distinctively, fuggin, mainly, unusually, so much, lower, fully, downright, total, a lot, clearly, consistently, strikingly, incredible, hella, exceedingly, extra, exponentially, unimaginable, clearest, super, kinda, true, tremendously, absolute, exceptionally, quintessentially, obviously, utter, only, low, massive, clear, fuckin, frickin, marginal, sortof, hardly, most, pretty, considerably, frigging, such, uber, sorta, collossal, lowest, totally, strongly, crucial, insanely, highly, passionately, monumental, partially, even, amazingly, major, almost, extremely, sort of, ridiculously, higher, big, largely, extraordinarily, terribly, enormously, less, mostly, more, decidedly, relatively, obvious, much, intensely, small, purely, wildly, monumentally, highest, mild, numerous, tremendous, deepest, certainly, inconsequential, some, kindof, friggin, double, serious, slight, considerable, scarce, profoundly, deeper, especially, quite a bit, definitely, remarkably, occasionally, fewest, minor, smaller, resounding, fairly, absolutely, immediately, dramatically, deep, effing, flipping, kind of, flippin, intensively, perfectly, moderately, high, outrageously, great, incredibly, extreme, significantly, slightest, way, vastly, marginally, entirely, endless, simply, abject, spectacularly, noticeably, clearer, several, a little, various, huge, infinite, unabashed, immediate, heckuva, truly, obscenely, less, multiple, exceptional, greatly, just enough, alot, partly, noticeable, rather, fabulously, fricking, thoroughly, immensely, particularly, universally, infinitely, stunningly, biggest, fucking, utterly, completely, unbelievably, important, few, occasional, mildly, smallest, substantially, frequently, visable, very, damn, constantly, enormous, slightly, pure, real, really, scarcely.

**List of negations:** not, nor, seldom, arent, won't, rarely, doesn't, barely, mustnt, hadn't, don't, isnt, nowhere, cannot, doesnt, couldn't, oughtnt, couldnt, haven't, never, ain't, despite, daren't, without, uhuh, scarcely, nope, wouldnt, neither, mustn't, nothing, needn't, dont, isn't, shouldn't, uh-uh, wasnt, neednt, werent, oughtn't, wouldn't, cant, nobody, hasn't, weren't, shan't, darent, shouldnt, aint, shant, hadnt, didnt, mightn't, none, wont, aren't, can't, didn't, 'mightnt', havent, wasn't, no, hardly, hasnt.

# ACKNOWLEDGEMENTS

We acknowledge the Rest-Mex Shared Task 2023 organizers *Álvarez-Carmona et al. (2023)* for providing us the possibility to evaluate the results in Table 9, given that the test dataset is not accessible to the general public.

# ADDITIONAL INFORMATION AND DECLARATIONS

## Funding

The European Research Council (ERC) funded this research under the Horizon Europe research and innovation programme (SALSA, grant agreement No. 101100615),

SCANNER-UDC (PID2020-113230RB-C21) funded by MICIU/AEI/10.13039/501100011033, LATCHING (PID2023-147129OB-C21) funded by MICIU/AEI/10.13039/501100011033 and ERDF (EU), project GAP (PID2022-139308OA-I00) funded by MICIU/AEI/10.13039/501100011033/ and ERDF (EU), Ministry for Digital Transformation and Civil Service and "NextGenerationEU" PRTR under grant TSI-100925-2023-1, Xunta de Galicia (ED431C 2024/02), and Galician Research Center "CITIC", funded by Xunta de Galicia through the collaboration agreement between the Consellería de Cultura, Educación, Formación Profesional e Universidades and the Galician universities for the reinforcement of the research centres of the Galician University System (CIGUS). Furthermore, this research was supported by the International, Interdisciplinary and Intersectoral Information and Communications Technology PhD programme (3-i ICT) granted to CITIC and supported by the European Union through the Horizon 2020 research and innovation programme under a Marie Skłodowska-Curie agreement (H2020-MSCA-COFUND), GA 101034261. The funders had no role in study design, data collection and analysis, decision to publish, or preparation of the manuscript.

## Grant Disclosures

The following grant information was disclosed by the authors:
European Research Council (ERC): 101100615.
MICIU/AEI/10.13039/501100011033: SCANNER-UDC (PID2020-113230RB-C21).
MICIU/AEI/10.13039/501100011033: LATCHING (PID2023-147129OB-C21).
ERDF (EU).
MICIU/AEI/10.13039/501100011033: GAP (PID2022-139308OA-I00).
Ministry for Digital Transformation and Civil Service and "NextGenerationEU" PRTR: TSI-100925-2023-1.
Xunta de Galicia: ED431C 2024/02.
Galician Research Center "CITIC": ED431C 2024/02.
Consellería de Cultura, Educación, Formación Profesional e Universidades.
Galician University System (CIGUS).
CITIC: 3-i ICT.
European Union: GA 101034261.

## Competing Interests

The authors declare that they have no competing interests.

## Author Contributions

- Muhammad Imran conceived and designed the experiments, performed the experiments, analyzed the data, performed the computation work, prepared figures and/or tables, authored or reviewed drafts of the article, and approved the final draft.
- Olga Kellert conceived and designed the experiments, performed the computation work, prepared figures and/or tables, authored or reviewed drafts of the article, and approved the final draft.

- Carlos Gómez-Rodríguez conceived and designed the experiments, prepared figures and/or tables, authored or reviewed drafts of the article, supervision, Funding acquisition, and approved the final draft.

## Data Availability

The following information was supplied regarding data availability:

The OpeNERen, OpeNERes, UD_English-EWT and UD_Spanish-AnCora datasets are available at Zenodo: Muhammad Imran. (2025). chimran135/syntax-injected-sentiment-analysis: Syntax-injected Sentiment Analysis (1.0.0). Zenodo. https://doi.org/10.5281/zenodo.15323755.

The Rest-Mex 2023 shared task website provides access to the training data for registered participants. The data are held privately for assessment purposes and can be obtained from Rest-Mex 2023 organizers https://sites.google.com/cimat.mx/rest-mex2023 by contacting Miguel Ángel Álvarez Carmona at miguel.alvarez@cimat.mx.

The Rest-Mex 2023 dataset was used under license for this study.